\documentclass[letterpaper]{article}
\usepackage{iccc}

\usepackage{algorithm}
\usepackage{algpseudocode}

\usepackage{graphicx}

\usepackage{xcolor}

\usepackage{times}
\usepackage{helvet}
\usepackage{courier}
\title{Creative Beam Search: LLM-as-a-Judge for Improving Response Generation}
\author{
Giorgio Franceschelli\\
Department of Computer Science and Engineering\\
Alma Mater Studiorum Università di Bologna\\
giorgio.franceschelli@unibo.it\\
\And Mirco Musolesi\\
Department of Computer Science\\
University College London\\
Department of Computer Science and Engineering\\
Alma Mater Studiorum Università di Bologna\\
m.musolesi@ucl.ac.uk\\
}
\begin{document} 
\maketitle
\begin{abstract}
\begin{quote}
Large language models are revolutionizing several areas, including artificial creativity. However, the process of generation in machines profoundly diverges from that observed in humans. In particular, machine generation is characterized by a lack of intentionality and an underlying creative process.
We propose a method called Creative Beam Search that uses Diverse Beam Search and LLM-as-a-Judge to perform response generation and response validation. The results of a qualitative experiment show how our approach can provide better output than standard sampling techniques. We also show that the response validation step is a necessary complement to the response generation step.
\end{quote}
\end{abstract}

\section{Introduction}

Recent advancements in deep learning have led to a wave of generative models, in particular large language models (LLMs), capable of impacting society at multiple levels \cite{bommasani2021opportunities}. Thanks to the quality of their outputs, the impact of LLMs on creative fields has been substantial \cite{newton2023ai,weidinger2022taxonomy}. However, LLMs are still far from being creative due to their lack of intentionality \cite{shanahan2024talking} and the absence of a genuinely creative process in their production \cite{franceschelli2023creativity}.

In this paper, we introduce Creative Beam Search (CBS), a novel generate-and-test sampling scheme designed to artificially replicate certain aspects of the creative process.
According to the framework proposed in \cite{amabile1983social}, creativity should involve the following steps: task presentation (from internal or external stimuli); preparation; response generation (thanks to creativity-relevant skills); and response validation (thanks to domain-relevant skills).
In particular, CBS first simulates the response generation phase through Diverse Beam Search (DBS) \cite{vijayakumar2018diverse}, generating a more diverse set of possible solutions. Then, it performs a \textit{self-evaluation phase} in LLM-as-a-Judge style \cite{zheng2023judging} to select the final output. We evaluate our method against the classic sampling strategy with a qualitative assessment study, finding that end-users find our approach preferable and, on average, CBS (as a generate-and-test approach with DBS) provides better solutions than DBS alone.

The remainder of the article is structured as follows. First, we review the relationship between LLMs and creativity and we introduce the key concepts at the basis of Creative Beam Search. Then, we detail our proposed method and present our qualitative experiment results. Finally, we discuss our findings and the limitations of the proposed approach, and we conclude with final remarks.

\section{Related Work} \label{relatedwork}

\subsection{LLMs and Creativity}
The potential impact of LLMs on creative fields has been evident since the advent of GPT models \cite{brown2020language,openai2023gpt4} and their competitors, e.g. \cite{touvron23llama}. Research has been conducted to determine whether LLMs can pass human creativity tests, such as the Alternate Uses Test \cite{stevenson2022putting}, and to explore ways to improve their results \cite{goes2023pushing}. However, their intrinsic lack of intentionality and consciousness should prevent them from being truly creative \cite{franceschelli2023creativity}.
Another area of research is focused on enhancing the ability of LLMs to generate creative outputs. For example, LLMs can be fine-tuned \cite{sawicki2023power} or used in zero-shot settings \cite{sawicki2023bits} to write in the style of famous authors. Another possibility is to use Reinforcement Learning from Human Feedback (RLHF) \cite{christiano2017deep} to teach an LLM to write haikus that human evaluators would find more creative \cite{pardinas2023leveraging}. Finally, active divergence techniques \cite{berns2020bridging} can also be used. Quality-diversity algorithms can help find more creative solutions by leveraging human feedback \cite{li2023quality} or AI feedback \cite{bradley2023quality} to measure quality.

\subsection{Beam Search}

Beam Search \cite{ott2018analyzing} is a text generation strategy that maintains several hypotheses (known as the beam budget $B$) at each time step and eventually chooses the hypothesis with the overall highest probability under the model. This approach, rather than focusing on single tokens (which can lead to sub-optimal or even degenerated solutions), considers the likelihood of the entire sequence \cite{caccia2020Language}. However, Beam Search often focuses on a single highly valued beam, resulting in final candidates that are merely minor variations of a single sequence. Diverse Beam Search \cite{vijayakumar2018diverse} proposes to overcome this issue by dividing the beam budget into $G$ groups. It enforces diversity between different groups by penalizing candidates that share tokens with other beams. This guarantees increased diversity in the final solutions. Other variants of Beam Search have been proposed as well, to enforce a certain constraint over the output \cite{hokamp2017lexically} or to substitute the likelihood with a self-evaluation scheme \cite{xie2023selfevaluation}.

\subsection{LLM-as-a-Judge}
The LLM-as-a-Judge approach involves the LLM evaluating its own responses. \cite{chiang2023large,zheng2023judging} show that evaluations from strong LLMs align with those from human experts. However, these evaluations suffer from positional bias, i.e., altering the order of candidate responses can affect their quality ranking \cite{wang2023large}. This new capability has led to the adoption of self-evaluation during training, replacing human feedback for RLHF \cite{bai2022constitutional,lee2023rlaif} or for other learning strategies \cite{chen2024selfplay,yuan2024selfrewarding}. In addition, LLM-as-a-Judge can be applied at inference time. It can guide quality-diversity search algorithms \cite{bradley2023quality} or improve responses for creativity tests \cite{goes2023pushing,summers2023brainstorm}.

\section{Creative Beam Search} \label{cbs}

\begin{figure}[h]
    \centering
    \includegraphics[width=.47\textwidth]{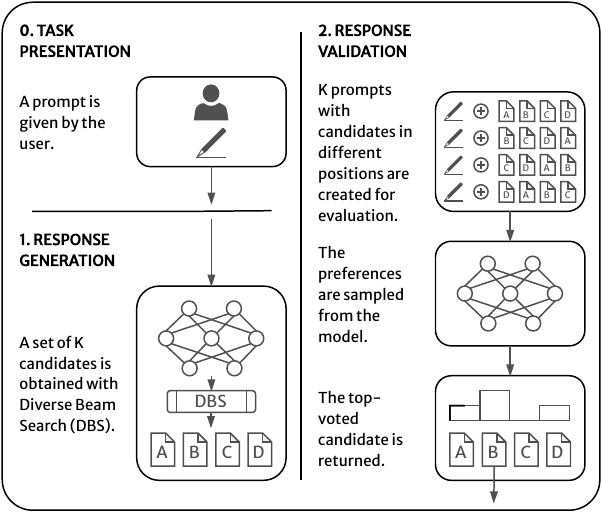}
    \caption{The Creative Beam Search method. Given a user prompt (step 0), DBS samples $K$ candidate solutions from a pre-trained language model (step 1). Then, $K$ evaluative prompts are composed by altering the order of the candidates and are passed to the model as inputs (step 2). The candidate with the most preferences is finally outputted.}
    \label{fig:cbs}
\end{figure}

Drawing from the componential model of creativity \cite{amabile1983social}, we propose a method, namely Creative Beam Search (CBS), to better simulate (parts of) the human creative process during text generation. In particular, after a task presentation step where an external stimulus is provided in the form of a user prompt and a preparation step where a pre-trained language model is loaded (bringing along the facts and information already acquired), CBS is articulated in two steps: response generation and response validation. The full process is summarized in Figure \ref{fig:cbs}.

\begin{figure*}[ht]
    \centering
    \includegraphics[width=.95\textwidth]{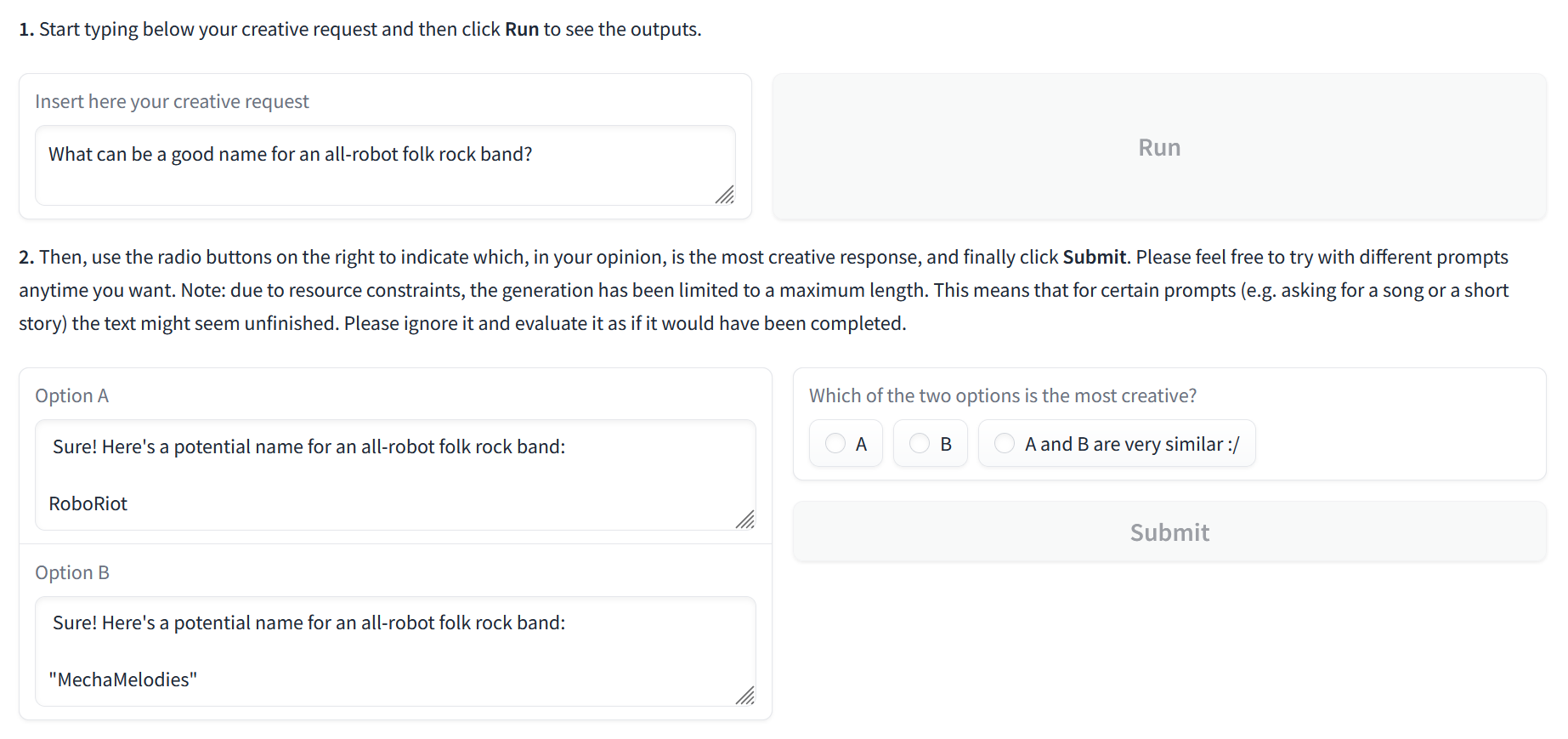}
    \caption{The interface presented to the end-users during our experiment. After inserting a prompt with a creative request, two options are shown in a random order: the CBS output and the standard sampling output. The user is then asked to indicate which is the most creative in their opinion (or if the two options are too similar to decide).}
    \label{fig:gradio}
\end{figure*}

\subsection{Response Generation} \label{responsegeneration}

During the response generation phase, an individual generates response possibilities by searching through the available pathways, exploring features that are relevant to the task at hand \cite{amabile1983social}. This process requires creativity-relevant skills as well as a method to limit the search to feasible and relevant solutions.

We propose to simulate these aspects using Diverse Beam Search for sequence generation. During beam search, a better collection of options is generated thanks to a diversity penalty. The beam budget $B$ is divided into $G$ groups. At each generation step, the $\frac{B}{G}$ solutions for a given group are selected among all possible $\frac{B}{G} \cdot |\mathcal{V}|$ candidates (where $\mathcal{V}$ is the vocabulary). These solutions optimize an objective consisting of two terms: the standard sequence likelihood under the model and a dissimilarity term that encourages diversity across groups. Commonly, Hamming diversity is considered, where each token receives a penalty proportional to the number of times that same token has been selected in other groups at the same step. Therefore, DBS can be seen as guided by two forces: the diversity penalty, which represents a simplified creativity-oriented skill, and the likelihood under the model, which helps focus the search to feasible and relevant paths.

\subsection{Response Validation} \label{responsevalidation}

During the response validation phase, the response possibilities are tested for quality and appropriateness, using the knowledge and assessment criteria from domain-relevant skills \cite{amabile1983social}. 

We propose an explicit self-assessment step that leverages the evaluative capabilities of recent generative models \cite{lee2023rlaif,yuan2024selfrewarding}. This involves asking the model to choose among the top $K$ candidates generated by DBS, according to their score. This allows the system to output the solution the model finds to be the best for the task, rather than simply returning the one with the highest combined likelihood and diversity. While \cite{amabile1983social} suggests evaluating a single response and repeating the entire process if the test is not passed, our method simplifies this by evaluating multiple candidates in a single step. This trade-off allows CBS to maintain short compute times, making it effective for online co-creative purposes.

In practice, CBS uses LLM-as-a-Judge prompting \cite{zheng2023judging} to make the model decide among the generated candidates. To address positional bias, we use the balanced position calibration scheme \cite{wang2023large}. We create $K$ different prompts by \textit{rotating} the top $K$ candidates, ensuring each candidate is considered in all possible positions. We then aggregate the votes and the candidate with the most preferences is selected. In the event of a tie, the initial order of the candidates (i.e., the DBS score) is taken into account. 



\section{Experiments} \label{experiments}

We conducted a qualitative evaluation of Creative Beam Search to assess its potential for co-creativity. Figure \ref{fig:gradio} shows a screenshot of the interface we used, which was created with Gradio \cite{abid2019gradio}.

\subsection{Setup}

We chose Llama 2 \cite{touvron23llama} as our pre-trained language model. Due to resource constraints, we selected the 7B variant and used the RLHF-tuned version, which provides more accurate and coherent responses. We set the beam budget $B$ to $8$, divided into single-item groups (i.e., $G=8$). The diversity penalty was scaled by a factor of $10$ to counterbalance the likelihood score. We then retained the top $K=4$ solutions for the evaluation step. For the DBS step, we used the prompt from Algorithm \ref{alg:gen_prompt}; the prompt for self-assessment is detailed in Algorithm \ref{alg:eval_prompt}.

\begin{algorithm}
\caption{Prompt for response generation.}\label{alg:gen_prompt}
\begin{algorithmic}
\State \{`role': `user', 
\State \hspace*{.5em}`content': `\textsc{\$input}. Provide only one answer without any explanation.'\}
\end{algorithmic}
\end{algorithm}

\begin{algorithm}
\caption{Prompt for response validation.}\label{alg:eval_prompt}
\begin{algorithmic}
\State \{`role': `user', 
\State \hspace*{.5em}`context': `Which of the following is the most creative answer to ``\textsc{\$input}''?
\State \hspace*{.5em}1) \textsc{\$candidate1}
\State \hspace*{.5em}2) \textsc{\$candidate2}
\State \hspace*{.5em}3) \textsc{\$candidate3}
\State \hspace*{.5em}4) \textsc{\$candidate4}
\State \hspace*{.5em}Provide only the number of the most creative answer without any explanation.'\}
\end{algorithmic}
\end{algorithm}

As mentioned above, we repeated the latter step $K=4$ times, each time altering the positions of the candidates.

We limited the model outputs to $256$ new tokens. Although this is a significant constraint, we believe it does not impact the final result as differences in creativity should be noticeable even in shorter texts. Lastly, we used a greedy decoding strategy (i.e., always selecting the most probable token) for the self-assessment to prevent the best candidate from being chosen randomly.

\subsection{Qualitative Results}

We carried out a qualitative evaluation involving 31 graduate students in Computer Science. They were given the freedom to input their prompts and were asked to choose between the CBS and the standard output (generated with a temperature of $1.0$ and nucleus sampling \cite{holtzman2020curious} with top-p of $0.9$). The presentation order of the two solutions was randomized, and the user could also indicate the outputs were too similar to differentiate. 

We gathered a total of 217 answers. As reported in Table \ref{tab_res}, CBS was preferred 45\% of the time, with a significant margin over the standard output. However, in about one-fourth of the cases, the responses were too similar to make a choice. This suggests that despite the diversity penalty and self-evaluation step, CBS output does not deviate significantly from standard sampling.

\begin{table}[t]
\centering
\begin{tabular}{|l c c c|} 
 \hline
 Preference & CBS != DBS & CBS == DBS & Total \\
 \hline
 CBS & \textbf{.34} & \textbf{.11} & \textbf{.45} \\ 
 STD & .18 & \textbf{.11} & .29 \\
 Same & .19 & .7 & .26 \\
 \hline
  & .71 & .29 & 1.00 \\
 \hline
\end{tabular}
\caption{Aggregate results from our qualitative assessment. The three possible preferences (CBS for Creative Beam Search, STD for standard sampling, and Same for when CBS and STD were too similar to choose) are divided considering whether CBS output is the same as Diverse Beam Search (DBS) output or not, and in total.\label{tab_res}}
\end{table}

We also tracked whether the candidate selected during self-evaluation was the same as the one selected by DBS. The overlap was 29\%, which is less than the 35.3\% that a random selection would have led to. This indicates that the self-evaluation step was not merely random and has subverted more than confirmed the DBS scoring.

\begin{figure}[t]
    \centering
    \includegraphics[width=.45\textwidth]{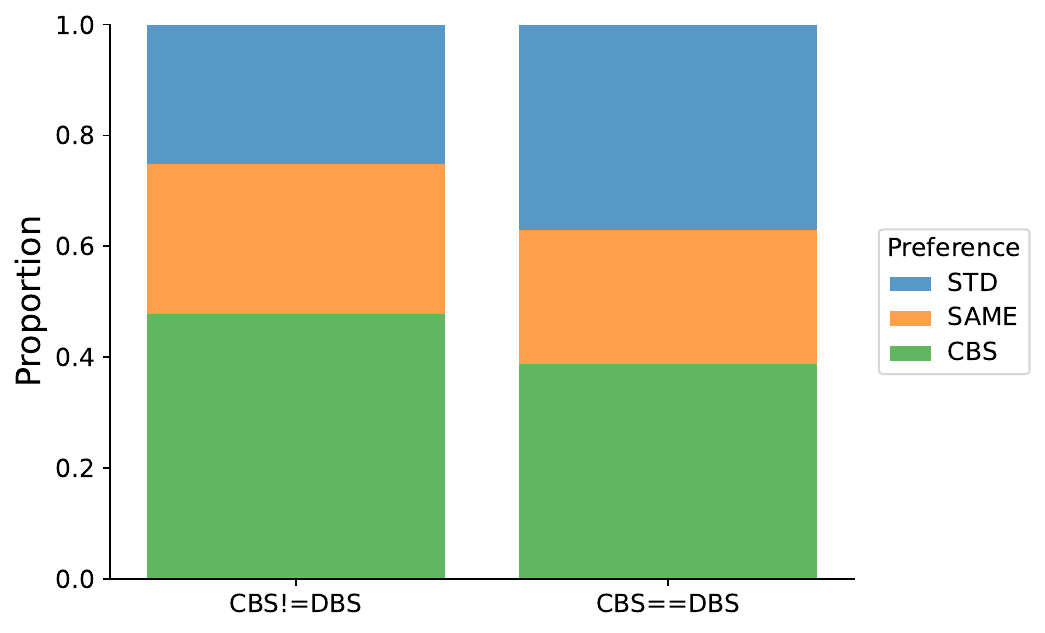}
    \caption{Percentage of end-users' preferences comparing when CBS output is equal to DBS output and when it is not.}
    \label{fig:cbs_dbs}
\end{figure}

Finally, we also analyze whether there was a difference in user preference for CBS outputs that matched or did not match the DBS outputs. Figure \ref{fig:cbs_dbs} shows the preference proportions for both scenarios. While the differences are not substantial, the standard output was preferred more when compared with the DBS output. This suggests that the final self-evaluation step can further improve Diverse Beam Search.

\section{Discussion} \label{discussion}

This paper has introduced a new sampling scheme, Creative Beam Search, to tackle the misalignment between the human creative process and how generative models produce their outputs. It leverages recent techniques such as Diverse Beam Search and LLM-as-a-Judge to simulate aspects of response generation and validation. However, it does not address other key aspects as outlined by \cite{amabile1983social}, such as task motivation from internal stimuli and the possibility of iteratively adjusting the responses. Moreover, both Diverse Beam Search and LLM-as-a-Judge have limitations. For instance, Diverse Beam Search uses Hamming diversity, which only considers differences at the same time step. This can lead to overly similar sequences due to minor misalignments such as initial spacing. In addition, it is only applicable to sequence generation tasks and is more expensive than classic decoding strategies. As for LLM-as-a-Judge, it is important to remark that LLMs are not conscious or intentional. Therefore, self-evaluation does not reflect any personal belief but merely returns what the model has learned to be more likely. Consequently, our approach can be considered as an artificial simulation of certain aspects of creativity. Finally, there is a need to extend the experimental evaluation, considering the impact of the prompt structure on the overall results.

Despite these limitations, our qualitative experiment shows that, on average, Creative Beam Search is viewed as a more creative sampling scheme than traditional methods by potential end-users. Furthermore, our results suggest that the self-evaluation step improves the output choice even when considering a small number of candidate solutions from DBS. Future work could explore whether considering a broader and more diverse set of candidates could lead to even better results. Thanks to its simplicity, our method can be easily extended to other, potentially more powerful, LLMs or to models trained with more creativity-oriented strategies. In conclusion, we believe our paper contributes to the growing field of generative learning for computational creativity \cite{franceschelli2024survey}.

\section{Acknowledgments}

The participation and presentation at ICCC'24 was supported by the ISA Doctoral Prize (ISA DP), offered by Istituto di Studi Avanzati, Alma Mater Studiorum Università di Bologna.

\bibliographystyle{iccc}
\bibliography{iccc}

\begin{thebibliography}{}

\bibitem[\protect\citeauthoryear{Abid \bgroup et al.\egroup }{2019}]{abid2019gradio}
Abid, A.; Abdalla, A.; Abid, A.; Khan, D.; Alfozan, A.; and Zou, J.
\newblock 2019.
\newblock Gradio: Hassle-free sharing and testing of {ML} models in the wild.
\newblock arXiv:1906.02569 [cs.LG].

\bibitem[\protect\citeauthoryear{Amabile}{1983}]{amabile1983social}
Amabile, T.~M.
\newblock 1983.
\newblock The social psychology of creativity: A componential conceptualization.
\newblock {\em Journal of Personality and Social Psychology} 45(2):357--376.

\bibitem[\protect\citeauthoryear{Bai \bgroup et al.\egroup }{2022}]{bai2022constitutional}
Bai, Y.; Kadavath, S.; Kundu, S.; Askell, A.; Kernion, J.; Jones, A.; Chen, A.; Goldie, A.; Mirhoseini, A.; McKinnon, C.; Chen, C.; Olsson, C.; Olah, C.; Hernandez, D.; Drain, D.; Ganguli, D.; Li, D.; Tran-Johnson, E.; Perez, E.; ...; and Kaplan, J.
\newblock 2022.
\newblock {Constitutional AI: Harmlessness from AI Feedback}.
\newblock arXiv:2212.08073 [cs.CL].

\bibitem[\protect\citeauthoryear{Berns and Colton}{2020}]{berns2020bridging}
Berns, S., and Colton, S.
\newblock 2020.
\newblock Bridging generative deep learning and computational creativity.
\newblock In {\em {Proceedings of the 11th International Conference on Computational Creativity (ICCC'20)}}.

\bibitem[\protect\citeauthoryear{Bommasani \bgroup et al.\egroup }{2021}]{bommasani2021opportunities}
Bommasani, R.; Hudson, D.; Adeli, E.; Altman, R.; Arora, S.; Arx, S.; Bernstein, M.; Bohg, J.; Bosselut, A.; Brunskill, E.; Brynjolfsson, E.; Buch, S.; Card, D.; Castellon, R.; Chatterji, N.; Chen, A.; Creel, K.; Davis, J.; Demszky, D.; ...; and Liang, P.
\newblock 2021.
\newblock On the opportunities and risks of foundation models.
\newblock arXiv:2108.07258 [cs.LG].

\bibitem[\protect\citeauthoryear{Bradley \bgroup et al.\egroup }{2023}]{bradley2023quality}
Bradley, H.; Dai, A.; Teufel, H.; Zhang, J.; Oostermeijer, K.; Bellagente, M.; Clune, J.; Stanley, K.; Schott, G.; and Lehman, J.
\newblock 2023.
\newblock Quality-{D}iversity through {AI} feedback.
\newblock arXiv:2310.13032 [cs.CL].

\bibitem[\protect\citeauthoryear{Brown \bgroup et al.\egroup }{2020}]{brown2020language}
Brown, T.; Mann, B.; Ryder, N.; Subbiah, M.; Kaplan, J.~D.; Dhariwal, P.; Neelakantan, A.; Shyam, P.; Sastry, G.; Askell, A.; Agarwal, S.; Herbert-Voss, A.; Krueger, G.; Henighan, T.; Child, R.; Ramesh, A.; Ziegler, D.; Wu, J.; Winter, C.; ...; and Amodei, D.
\newblock 2020.
\newblock Language models are few-shot learners.
\newblock In {\em Advances in Neural Information Processing Systems ({NIPS'20})}.

\bibitem[\protect\citeauthoryear{Caccia \bgroup et al.\egroup }{2020}]{caccia2020Language}
Caccia, M.; Caccia, L.; Fedus, W.; Larochelle, H.; Pineau, J.; and Charlin, L.
\newblock 2020.
\newblock Language {GANs} falling short.
\newblock In {\em {Proceedings of the 8th International Conference on Learning Representations (ICLR'20)}}.

\bibitem[\protect\citeauthoryear{Chen \bgroup et al.\egroup }{2024}]{chen2024selfplay}
Chen, Z.; Deng, Y.; Yuan, H.; Ji, K.; and Gu, Q.
\newblock 2024.
\newblock Self-play fine-tuning converts weak language models to strong language models.
\newblock arXiv:2401.01335 [cs.LG].

\bibitem[\protect\citeauthoryear{Chiang and Lee}{2023}]{chiang2023large}
Chiang, C.-H., and Lee, H.-y.
\newblock 2023.
\newblock Can large language models be an alternative to human evaluations?
\newblock In {\em {Proceedings of the 61st Annual Meeting of the Association for Computational Linguistics (ACL'23)}}.

\bibitem[\protect\citeauthoryear{Christiano \bgroup et al.\egroup }{2017}]{christiano2017deep}
Christiano, P.~F.; Leike, J.; Brown, T.; Martic, M.; Legg, S.; and Amodei, D.
\newblock 2017.
\newblock Deep reinforcement learning from human preferences.
\newblock In {\em {Advances in Neural Information Processing Systems (NeurIPS'17)}}.

\bibitem[\protect\citeauthoryear{Ding \bgroup et al.\egroup }{2023}]{li2023quality}
Ding, L.; Zhang, J.; Clune, J.; Spector, L.; and Lehman, J.
\newblock 2023.
\newblock Quality diversity through human feedback.
\newblock In {\em {Proceedings of the NeurIPS'23 ALOE Workshop}}.

\bibitem[\protect\citeauthoryear{Franceschelli and Musolesi}{2023}]{franceschelli2023creativity}
Franceschelli, G., and Musolesi, M.
\newblock 2023.
\newblock On the creativity of large language models.
\newblock arXiv:2304.00008 [cs.AI].

\bibitem[\protect\citeauthoryear{Franceschelli and Musolesi}{2024}]{franceschelli2024survey}
Franceschelli, G., and Musolesi, M.
\newblock 2024.
\newblock Creativity and machine learning.
\newblock {\em ACM Computing Surveys}.
\newblock Accepted for Publication. To Appear.

\bibitem[\protect\citeauthoryear{Goes \bgroup et al.\egroup }{2023}]{goes2023pushing}
Goes, F.; Volpe, M.; Sawicki, P.; Grzés, M.; and Watson, J.
\newblock 2023.
\newblock Pushing {GPT}’s creativity to its limits: {A}lternative {U}ses and {T}orrance {T}ests.
\newblock In {\em {Proceedings of the 14th International Conference on Computational Creativity (ICCC'23)}}.

\bibitem[\protect\citeauthoryear{Hokamp and Liu}{2017}]{hokamp2017lexically}
Hokamp, C., and Liu, Q.
\newblock 2017.
\newblock Lexically constrained decoding for sequence generation using grid beam search.
\newblock In {\em {Proceedings of the 55th Annual Meeting of the Association for Computational Linguistics (ACL'17)}}.

\bibitem[\protect\citeauthoryear{Holtzman \bgroup et al.\egroup }{2020}]{holtzman2020curious}
Holtzman, A.; Buys, J.; Du, L.; Forbes, M.; and Choi, Y.
\newblock 2020.
\newblock The curious case of neural text degeneration.
\newblock In {\em {Proceedings of the 8th International Conference on Learning Representations (ICLR'20)}}.

\bibitem[\protect\citeauthoryear{Lee \bgroup et al.\egroup }{2023}]{lee2023rlaif}
Lee, H.; Phatale, S.; Mansoor, H.; Lu, K.; Mesnard, T.; Bishop, C.; Carbune, V.; and Rastogi, A.
\newblock 2023.
\newblock {RLAIF}: Scaling reinforcement learning from human feedback with {AI} feedback.
\newblock arXiv:2309.00267 [cs.CL].

\bibitem[\protect\citeauthoryear{Newton and Dhole}{2023}]{newton2023ai}
Newton, A., and Dhole, K.
\newblock 2023.
\newblock Is {AI} art another industrial revolution in the making?
\newblock In {\em {Proceedings of the AAAI'23 Creative AI Across Modalities Workshop}}.

\bibitem[\protect\citeauthoryear{OpenAI}{2023}]{openai2023gpt4}
OpenAI.
\newblock 2023.
\newblock {GPT-4 Technical Report}.
\newblock arXiv:2303.08774 [cs.CL].

\bibitem[\protect\citeauthoryear{Ott \bgroup et al.\egroup }{2018}]{ott2018analyzing}
Ott, M.; Auli, M.; Grangier, D.; and Ranzato, M.
\newblock 2018.
\newblock Analyzing uncertainty in neural machine translation.
\newblock In {\em {Proceedings of the 35th International Conference on Machine Learning (ICML'18)}}.

\bibitem[\protect\citeauthoryear{Pardinas \bgroup et al.\egroup }{2023}]{pardinas2023leveraging}
Pardinas, R.; Huang, G.; Vazquez, D.; and Pich{\'e}, A.
\newblock 2023.
\newblock Leveraging human preferences to master poetry.
\newblock In {\em {Proceedings of the AAAI'23 Workshop on Creative AI Across Modalities}}.

\bibitem[\protect\citeauthoryear{Sawicki \bgroup et al.\egroup }{2023a}]{sawicki2023bits}
Sawicki, P.; Grzés, M.; Goes, F.; Brown, D.; Peeperkorn, M.; Khatun, A.; and Paraskevopoulou, S.
\newblock 2023a.
\newblock Bits of {G}rass: Does {GPT} already know how to write like {W}hitman?
\newblock In {\em {Proceedings of the 14th International Conference on Computational Creativity (ICCC'23)}}.

\bibitem[\protect\citeauthoryear{Sawicki \bgroup et al.\egroup }{2023b}]{sawicki2023power}
Sawicki, P.; Grzés, M.; Goes, F.; Brown, D.; Peeperkorn, M.; Khatun, A.; and Paraskevopoulou, S.
\newblock 2023b.
\newblock On the power of special-purpose {GPT} models to create and evaluate new poetry in old styles.
\newblock In {\em {Proc. of the 14th International Conference on Computational Creativity (ICCC'23)}}.

\bibitem[\protect\citeauthoryear{Shanahan}{2024}]{shanahan2024talking}
Shanahan, M.
\newblock 2024.
\newblock Talking about large language models.
\newblock {\em Communications of the ACM} 67(2):68–79.

\bibitem[\protect\citeauthoryear{Stevenson \bgroup et al.\egroup }{2022}]{stevenson2022putting}
Stevenson, C.; Smal, I.; Baas, M.; Grasman, R.; and van~der Maas, H.
\newblock 2022.
\newblock Putting {GPT-3}'s creativity to the ({A}lternative {U}ses) {T}est.
\newblock In {\em {Proceedings of the 13th International Conference on Computational Creativity (ICCC'22})}.

\bibitem[\protect\citeauthoryear{Summers-Stay, Voss, and Lukin}{2023}]{summers2023brainstorm}
Summers-Stay, D.; Voss, C.~R.; and Lukin, S.~M.
\newblock 2023.
\newblock Brainstorm, then select: a generative language model improves its creativity score.
\newblock In {\em {Proceedings of the AAAI'23 Workshop on Creative AI Across Modalities}}.

\bibitem[\protect\citeauthoryear{Touvron \bgroup et al.\egroup }{2023}]{touvron23llama}
Touvron, H.; Martin, L.; Stone, K.; Albert, P.; Almahairi, A.; Babaei, Y.; Bashlykov, N.; Batra, S.; Bhargava, P.; Bhosale, S.; Bikel, D.; Blecher, L.; Ferrer, C.~C.; Chen, M.; Cucurull, G.; Esiobu, D.; Fernandes, J.; Fu, J.; Fu, W.; ...; and Scialom, T.
\newblock 2023.
\newblock Llama 2: Open foundation and fine-tuned chat models.
\newblock arXiv:2307.09288 [cs.CL].

\bibitem[\protect\citeauthoryear{Vijayakumar \bgroup et al.\egroup }{2018}]{vijayakumar2018diverse}
Vijayakumar, A.; Cogswell, M.; Selvaraju, R.; Sun, Q.; Lee, S.; Crandall, D.; and Batra, D.
\newblock 2018.
\newblock Diverse beam search for improved description of complex scenes.
\newblock {\em {Proceedings of the 32nd AAAI Conference on Artificial Intelligence (AAAI'18)}}.

\bibitem[\protect\citeauthoryear{Wang \bgroup et al.\egroup }{2023}]{wang2023large}
Wang, P.; Li, L.; Chen, L.; Cai, Z.; Zhu, D.; Lin, B.; Cao, Y.; Liu, Q.; Liu, T.; and Sui, Z.
\newblock 2023.
\newblock Large language models are not fair evaluators.
\newblock arXiv:2305.17926 [cs.CL].

\bibitem[\protect\citeauthoryear{Weidinger \bgroup et al.\egroup }{2022}]{weidinger2022taxonomy}
Weidinger, L.; Uesato, J.; Rauh, M.; Griffin, C.; Huang, P.-S.; Mellor, J.; Glaese, A.; Cheng, M.; Balle, B.; Kasirzadeh, A.; Biles, C.; Brown, S.; Kenton, Z.; Hawkins, W.; Stepleton, T.; Birhane, A.; Hendricks, L.~A.; Rimell, L.; Isaac, W.; ...; and Gabriel, I.
\newblock 2022.
\newblock Taxonomy of risks posed by language models.
\newblock In {\em {Proceedings of the 2022 ACM Conference on Fairness, Accountability, and Transparency (FAccT'22)}}.

\bibitem[\protect\citeauthoryear{Xie \bgroup et al.\egroup }{2023}]{xie2023selfevaluation}
Xie, Y.; Kawaguchi, K.; Zhao, Y.; Zhao, X.; Kan, M.-Y.; He, J.; and Xie, Q.
\newblock 2023.
\newblock Self-evaluation guided beam search for reasoning.
\newblock In {\em {Proceedings of the 37th Conference on Neural Information Processing Systems (NIPS'23)}}.

\bibitem[\protect\citeauthoryear{Yuan \bgroup et al.\egroup }{2024}]{yuan2024selfrewarding}
Yuan, W.; Pang, R.~Y.; Cho, K.; Li, X.; Sukhbaatar, S.; Xu, J.; and Weston, J.
\newblock 2024.
\newblock Self-rewarding language models.
\newblock arXiv:2401.10020 [cs.CL].

\bibitem[\protect\citeauthoryear{Zheng \bgroup et al.\egroup }{2023}]{zheng2023judging}
Zheng, L.; Chiang, W.-L.; Sheng, Y.; Zhuang, S.; Wu, Z.; Zhuang, Y.; Lin, Z.; Li, Z.; Li, D.; Xing, E.; Zhang, H.; Gonzalez, J.~E.; and Stoica, I.
\newblock 2023.
\newblock Judging {LLM}-as-a-judge with {MT}-bench and chatbot arena.
\newblock In {\em {Proceedings of the 37th Conference on Neural Information Processing Systems Datasets and Benchmarks Track (NIPS'23)}}.

\end{thebibliography}

\end{document}